\documentclass[10pt,twocolumn,letterpaper]{article}

\usepackage[pagenumbers]{cvpr} 

\usepackage{graphicx}
\usepackage{amsmath}
\usepackage{amssymb}
\usepackage{booktabs}

\usepackage{multirow}
\usepackage{paralist}

\usepackage{comment}
\newcommand{\tabincell}[2]{\begin{tabular}{@{}#1@{}}#2\end{tabular}}
\usepackage[pagebackref=true,breaklinks=true,colorlinks,citecolor=blue,linkcolor=blue,bookmarks=false]{hyperref}

\usepackage[capitalize]{cleveref}
\crefname{section}{Sec.}{Secs.}
\Crefname{section}{Section}{Sections}
\Crefname{table}{Table}{Tables}
\crefname{table}{Tab.}{Tabs.}

\def\cvprPaperID{2540} 
\def\confName{CVPR}
\def\confYear{2022}

\begin{document}

\title{An Informative Tracking Benchmark}

\author{
    Xin Li$^{1}$
    \hspace{8pt}
    Qiao Liu$^{2}$
    \hspace{8pt}
    Wenjie Pei$^{3}$
    \hspace{8pt}
    Qiuhong Shen$^{3}$
    \hspace{8pt}
    Yaowei Wang$^{1}$\\
    Huchuan Lu$^{4,1}$
    \hspace{8pt}
    Ming-Hsuan Yang$^{5,6,7}$\\
    $^1$Peng Cheng Laboratory\hspace{12pt}
    $^2$Chongqing Normal University\\
    $^3$Harbin Institute of Technology, Shenzhen\hspace{12pt}
    $^4$Dalian University of Technology\\
    $^5$Google Research\hspace{12pt}
    $^6$University of California, Merced\hspace{12pt}
    $^7$Yonsei University\hspace{12pt}
}

\maketitle

\begin{abstract}
Along with the rapid progress of visual tracking, existing benchmarks become less informative due to redundancy of samples and weak discrimination between current trackers, making evaluations on all datasets extremely time-consuming.
Thus, a small and informative benchmark, which covers all typical challenging scenarios to facilitate assessing the tracker performance, is of great interest.
%
In this work, we develop a principled way to construct a small and informative tracking benchmark (ITB)
with 7\% out of 1.2 M frames of existing and newly collected datasets, which enables efficient evaluation while ensuring effectiveness.
%
Specifically, we first design a quality assessment mechanism to select the most informative sequences from existing benchmarks taking into account 1) challenging level, 2) discriminative strength, 3) and density of appearance variations.
%
%
%
Furthermore, we collect additional sequences to ensure the diversity and balance of tracking scenarios, leading to a total of 20 sequences for each scenario.
By analyzing the results of 15 state-of-the-art trackers re-trained on the same data, we determine the effective methods for robust tracking under each scenario and demonstrate new challenges for future research direction in this field.
The database and evaluation toolkit can be found at \url{https://github.com/XinLi-zn/Informative-tracking-benchmark}.

\end{abstract}

\vspace{-4mm}
\section{Introduction}
\label{sec:intro}
Visual tracking aims to estimate the states (\ie, location and extent) of
a target object in images based on an initialized target location.
It is a fundamental task in computer vision with a wide range of applications, such as intelligent surveillance, human-computer interaction, and video analysis.
With the advances of deep learning, significant progress has been made and certain challenging attributes~\cite{OTB2013,OTB2015} have been well studied, such as illumination variation, scale variation, and deformation.
However, tracking performance under real application scenarios where diverse challenging factors usually occur simultaneously, has not been explored extensively.
Therefore, it is crucial to evaluate the state-of-the-art deep trackers under diverse scenarios and identify new challenges to help design more robust methods.

\begin{figure}
    \centering
    \includegraphics[width=0.96\linewidth]{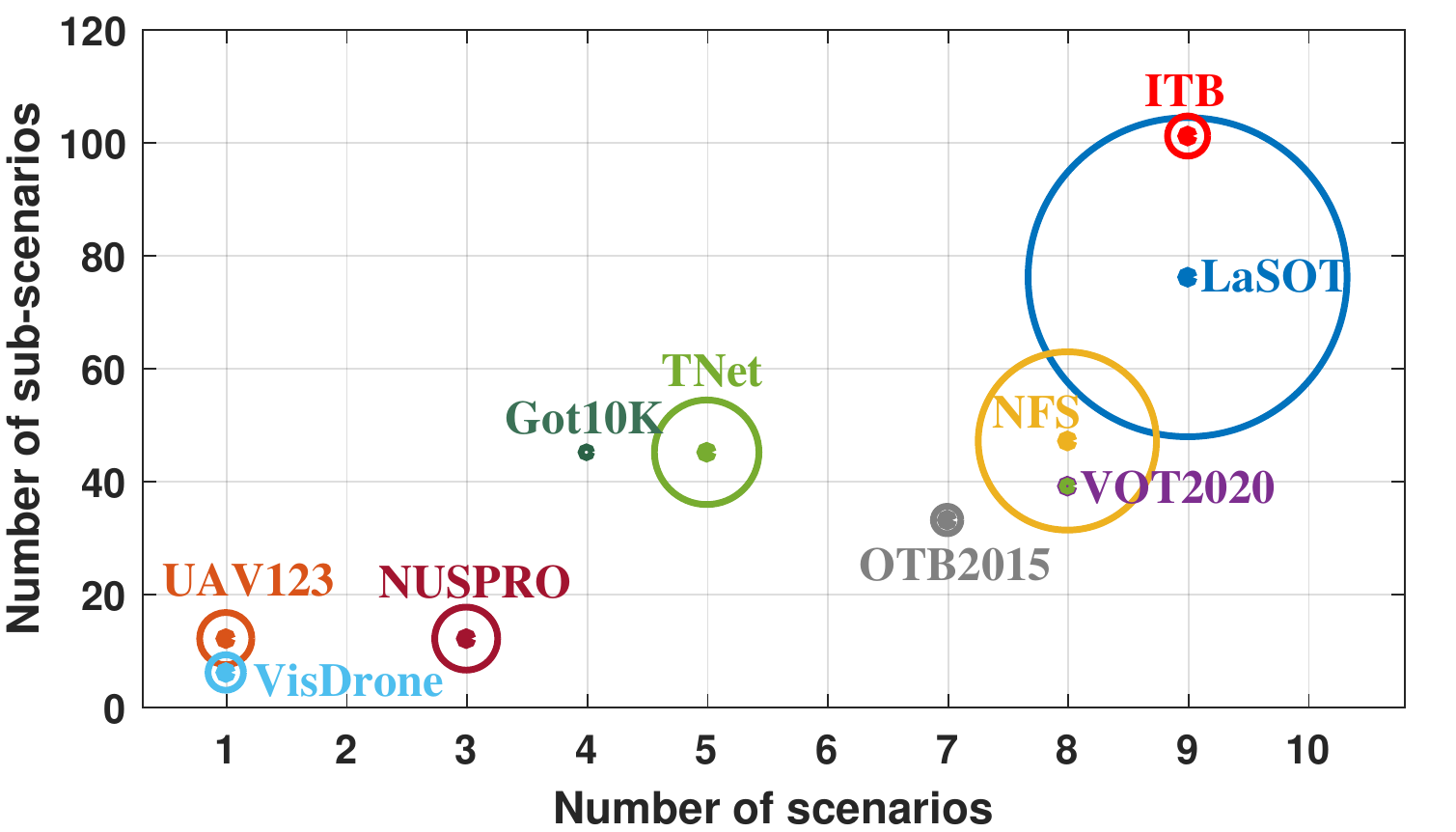}
    \vspace{-1mm}
    \caption{\textbf{Comparison of existing tracking benchmarks in terms of diversity and evaluation time.} The benchmarks include OTB2015~\cite{OTB2015}, UAV123~\cite{UAV123}, NFS~\cite{NFS}, NUSPRO~\cite{NUS-PRO}, GOT10K~\cite{GOT10k}, VisDrone~\cite{VisDrone}, VOT2020~\cite{VOT2020}, TNet~\cite{TrackingNet}, LaSOT~\cite{LaSOT}, and the proposed ITB.
    The circle size is proportional to the test time on the corresponding dataset.
    It shows that the proposed ITB benchmark includes the most diverse scenarios and sub-scenarios, while enabling efficient performance evaluation.
    }
    \label{fig:motivation}
    \vspace{-4mm}
\end{figure}

Numerous tracking benchmarks have been proposed to evaluate tracking performance mainly with single challenging attributes~\cite{OTB2015}, such as occlusion, scale variation, and fast motion, or a specific scenario, such as sequences captured from a UAV view~\cite{UAV123,VisDrone}, with higher frame rates~\cite{NFS}, long-term duration~\cite{LaSOT}, and specific object categories~\cite{NUS-PRO}.
%
However, the rapid advances of visual tracking necessitate the development of new benchmark that considers several factors.
%
First, existing attributes defined independently are less effective to reflect challenges in real applications which usually involve complicated combinations of multiple factors.
In addition, the scenarios in existing benchmarks either focus on target category or motion pattern, which are not sufficient to represent challenges in real applications.
%
Second, as significant progress has been made by recent tracking methods, there remains less room for in-depth performance evaluation.
On the other hand, some benchmarks contain long sequences but with almost static scenes with few appearance variations.
This makes the evaluation on {\em all} benchmarks extremely time-consuming and less effective in distinguishing the performance of different methods.
Third, although the codes of most existing trackers are released, the reported results are usually obtained using different training datasets, and certain hyper-parameters are finetuned in the testing stage.
Although evaluated on the same benchmarks, such approaches fail to clearly demonstrate how a tracking method performs.

In this paper, we construct an informative benchmark with diverse scenarios and high-quality sequences to address the above-mentioned issues.
We first propose a metric to measure the quality of sequences for effective evaluation by taking challenge level, discriminative strength, and density of appearance variations into consideration.
%
%
Based on the metric, we select the most informative/high-quality sequences from existing benchmarks such that the performance ranking remains the same.
To ensure diversity and balance between different scenarios, we collect additional sequences with a high density of target variations to cover 9 scenarios and ensure 20 sequences for each scenario.
To better reflect tracking challenges in real applications, we define the scenarios with subcategories by considering both appearance and variation patterns of target objects.
%
The informative benchmark constructed with 7\% out of 1.2 M frames allows saving 93\% of evaluation time (3,625 seconds on informative benchmark \vs 50,000 seconds on all benchmarks) for a real-time tracker (24 frames per second).
%
We re-train 15 state-of-the-art methods on the same dataset, and evaluate them on the proposed benchmark using a unified protocol.
%
%
In addition, we discuss new tracking challenges and potential research directions in this field.
%
We make the following contributions in this work:
\begin{compactitem}
\item We propose a quality metric to measure the effectiveness of a sequence for performance evaluation.
Based on this metric, we collect high-quality sequences that are challenging, discriminative, and with a high density of variations.
    %
\item We construct an informative tracking benchmark with 9 commonly seen scenarios and diverse sub-scenarios from widely-used datasets for efficient and effective performance evaluation.
The proposed benchmark demonstrates the performance of state-of-the-art trackers on various scenarios closer to real applications.
%
\item We develop an evaluation protocol using specified training and validation settings to facilitate fair comparisons of deep trackers.
We re-train 15 representative deep trackers using this protocol and evaluate them on the proposed benchmark for informative and fair assessment.
In addition, we discuss new challenges for future research work on object tracking.
\end{compactitem}

\section{Related Work}
\label{sec:related-work}
In this section, we review and discuss recent deep tracking methods and existing tracking benchmarks to give a brief summary of the current status of visual tracking.

\vspace{1mm}
\noindent\textbf{Deep tracking methods.}
Recently, visual tracking achieves significant progress along with the development of deep learning models that provide better representations and robust predictions.
Based on the type of the backbone models, deep trackers can be roughly divided into Siamese-based trackers~\cite{SIAMESEFC, DaSiamRPN,SiamRPN, C-RPN, SiamBAN, SiamR-CNN}, classifier-IoUNet tracking frameworks~\cite{ATOM,DiMP, PrDiMP, KYS}, transformer-based approaches~\cite{TrDiMP, TransT, Stark}, and other representative methods~\cite{DSLT,Meta-tracker,USOT,STMT}.

The Siamese framework casts tracking as a similarity learning problem, which first computes similarity feature maps between the target exemplar and the search region patch, and then estimates the target state based on the similarity maps.
Since the similarity map is independent of target category information, the Siamese framework can be trained offline using data with casual classes and used for tracking arbitrary targets without online learning or adaptation.
After that, numerous feature backbones and prediction modules have been introduced into the Siamese framework for achieving more robust and accurate performance.
For instance, SiamRPN~\cite{SiamRPN} builds an RPN module~\cite{fasterRCNN} upon the similarity feature maps, SiamRPN++~\cite{SiamRPN++} uses a deeper backbone network (ResNet50~\cite{RES50}) and a depth-wise correlation, SiamGAT~\cite{SiamGAT} adopts a graph attention network for better representation, and OCEAN~\cite{OCEAN} explores an anchor-free predictor for improving precision.
Benefiting from the offline training on large-scale data, Siamese-based methods achieve accurate performance.
However, they are less effective in handling long-term variations and distractors due to the lack of online adaption.


Another type of deep trackers~\cite{ATOM} (denoted as classifier-IoUNet) combine the discriminative ability of online learned classifiers and general priors about object boundary of an offline trained IoUNet~\cite{IoUNet} for enabling robust and accurate tracking performance.
To further improve the classifier module with offline training, Bhat \etal ~\cite{DiMP} propose a model prediction architecture, which exploits both target and background appearance information for predicting a classifier model in a meta-learning manner.
In addition, Danelljan \etal ~\cite{PrDiMP} introduce a probabilistic regression formulation to both the classifier and IoUNet components for better modeling the target state with label noise and ambiguities.
The classifier-IoUNet tracking framework achieves robust performance with the aid of online learning.
However, its performance is not stable due to the random factors in online training and the IoU module has difficulty in predicting target states with inconspicuous boundaries.

Motivated by the success of Transformer~\cite{transformer} in computer vision, such as visual recognition~\cite{VIT} and object detection~\cite{DETR}, several transformer-based trackers~\cite{TransT,TrDiMP,Stark} are proposed.
To exploit the powerful representation ability, Wang \etal ~\cite{TrDiMP} apply the transformer architecture to seek rich temporal information from video frames, while Chen \etal ~\cite{TransT} develop a transformer-based feature fusion network to generate better semantic feature maps between target exemplar and search region.
In addition, Xue \etal \cite{Stark} introduce an end-to-end transformer-based tracking framework, eliminating post-processing steps like box smoothing and cosine window weighting.
Benefited from better representation ability, the transformer-based approaches set new state-of-the-art performance on existing datasets.
However, visual tracking still faces numerous challenges and is far from real applications, which we will study in this paper.

\vspace{1mm}
\noindent\textbf{Tracking benchmarks.}
Numerous tracking benchmarks have been developed with increasing scales and diverse challenges for evaluating trackers.
The OTB benchmark~\cite{OTB2013,OTB2015} first introduces a tracking dataset with 100 sequences tagged with various attributes, and the metrics of precision and success plots, which are adopted by most benchmarks.
To evaluate model robustness, the VOT challenge~\cite{VOT2014} uses a re-initialization scheme after tracking failure and counts the failure times for computing the robustness score.
The OTB and VOT datasets are mainly composed of short-term sequences with various changes, while other existing datasets mainly focus on evaluating trackers under specific kinds of scenarios.
For instance, the TC128 dataset~\cite{TC128} comprising 128 RGB videos, investigates the effect of color information for visual tracking, the NFS dataset~\cite{NFS} with high frame rate sequences, studies tracking objects with fast motion, the NUS-PRO dataset~\cite{NUS-PRO} focuses on pedestrians and rigid objects, and the UAV123~\cite{UAV123} and VisDrone~\cite{VisDrone} datasets evaluate tracking methods under the UAV view.
Recently, large-scale datasets are developed for evaluating tracking performance more thoroughly.
The TNet benchmark~\cite{TrackingNet} contains 511 test videos with diverse object classes and the LaSOT benchmark~\cite{LaSOT} includes 280 long-term sequences with 701 K test frames in total.
In addition, the GOT-10k benchmark~\cite{GOT10k} collects test sequences with rich motion patterns for assessing tracking performance in the wild with complicated target movements.

With the significant advances in tracking methods, most sequences in existing benchmarks are becoming saturated and some attributes have been well addressed in less complicated scenarios.
However, it is still far from effective application in real scenarios.
In contrast, we propose a small benchmark with diverse scenarios defined by typical appearance and changing patterns for better evaluating how state-of-the-art methods perform in various types of scenarios closer to real applications.

\begin{table}[t]
\begin{center}
\caption{\textbf{Evaluated deep tracking methods.} The table shows 15 representative trackers using different base frameworks. For learning schemes, on: online learning, off: offline learning. The speed is tested on a PC with an RTX2080 GPU.}
\label{Tab:methods}
\vspace{-2mm}
\scriptsize
\renewcommand\arraystretch{1}
\resizebox{0.98\linewidth}{!}{
\begin{tabular}{l l c c c c }
\toprule
Framework & Method                  &  \tabincell{c}{Learning\\scheme}  & \tabincell{c}{Speed\\fps} &  \tabincell{c}{Publish\\year}  \\
\midrule
\multirow{5}{*}{Siamese}&RPN~\cite{SiamRPN}      &   off       &   80    &  2018 \\
                        &TADT~\cite{TADT}        &    on       &   38    &  2019 \\
                        &RPN++~\cite{SiamRPN++}  &     off     &   56   & 2019 \\ 
                        &Ocean~\cite{OCEAN}      &    off      &   24   &   2020  \\   
                        &SiamGAT~\cite{SiamGAT}  &      off    &   32 &  2021 \\    
\midrule  
\multirow{4}{*}{\tabincell{c}{Classifier\\+ IoUNet}}&ATOM~\cite{ATOM}       &  off+on     &  31    &  2019\\
                                                    &DiMP~\cite{DiMP}       &    off+on   &  54  &  2019\\
                                                    &PrDiMP~\cite{PrDiMP}    &   off+on   &  32   &   2020\\
                                                    &KYS~\cite{KYS}          &   off+on   &   21   &   2020\\            
\midrule                                                
\multirow{3}{*}{Transformer}&TransT~\cite{TransT}    &   offline     &   35   &  2021  \\
                            &TrDiMP~\cite{TrDiMP}    &      off+on   &   32   &  2021  \\
                            &Stark101~\cite{Stark}      &       off     &   33    &     2021\\ 

\midrule
\multirow{3}{*}{Others}&ECO~\cite{ECO}        &   on            &      62      &        2017 \\
                        &UDT~\cite{UDT}       &    off+on       &    70   &   2019 \\
                        &LTMU~\cite{LTMU}      &   off+on          &   14     &    2020\\
\bottomrule
\end{tabular}}
\end{center}
\vspace{-9mm}
\end{table}

\begin{table*}[t]
\begin{center}
\caption{\textbf{Comparison of ITB against existing benchmarks.} The table shows the numbers of the sequences belonging to each scenario and number of sub-scenarios of every dataset. In addition, the right part gives the global statistics. It shows that the proposed ITB covers all scenarios evenly and contains the most sub-scenarios with a number of 101. }
\label{Tab:datasets}
\vspace{-2mm}
\scriptsize
\renewcommand\arraystretch{1.1}
\setlength\tabcolsep{4pt}
 \resizebox{0.98\linewidth}{!}{
\begin{tabular}{l| ccccccccccc r}
\toprule
    \multicolumn{1}{r|}{Info.}        & \multicolumn{9}{c|}{Scenarios (video num. / sub-scenario num.)} & \multirow{2}{*}{\tabincell{c}{sub-scen.\\Num.} }&\multirow{2}{*}{\tabincell{c}{video\\Num.} }&\multirow{2}{*}{\tabincell{c}{Total\\frames} }\\
Dataset            & Human body & Human part & Animal  & Vehicle & Sign and logo & Sport ball & 3D-object   &   UAV     & Cartoon &   &   & \\ \midrule
            
OTB2015    & 36 / 10 & 26 / 3& 3 / 3 & 14 / 3 &  3 / 3  & -- & 16 / 10&  --  & 2 / 1 & 33 & 100 & 59 K \\
VOT-2020   & 15 / 8  &  12 / 3 & 14 / 11 &  4 / 4 & 7 / 7 & 3 / 3  & 2 / 2&3 / 1 & --  & 39  & 60 & 21 K \\
NUS-PRO    & 205 / 6&   60 / 1& -- &  100 / 5&   -- &  --  &  --   &   --   &  -- & 12  & 365  & 135 K \\
UAV123     &  --&   -- &  -- &    -- &  -- &  -- &   -- &  123 / 12 &  -- & 12  & 123  &  113 K     \\
VisDrone   &  --  &   -- &  --&  --  &  -- &  -- &   --  &  97 / 6 &  -- & 6  & 97  &  76 K      \\
NFS(240)   & 20 / 11 &  6 / 4& 14 / 7  &  15 / 6 &  5 / 5&  22 / 6 &  17 / 7 &   1 / 1 & --  & 47  & 100    &   383 K    \\
TNet&  131 / 11& 4 / 1 & 151 / 18 &  158 / 9 &   --  &   --& 65 / 6 &  -- & -- &  45 & 511    &  225 K\\
Got-10K    & 30 / 7  &  --& 109 / 28& 22 / 5&--  &  -- &  19 / 5 &   -- & -- & 45  & 180    & 23 K      \\ 
LaSOT      & 4 / 4& 12 / 3 & 136 / 34  & 40 / 10 & 4 / 1& 12 /3  &  64 / 16 & 4 / 1& 4 / 4 & 76  & 280    & 685 K  \\\midrule
ITB    &  20 / 17 & 20 / 7 & 20 / 15 & 20 / 13 & 20 / 6  &  20 / 8  & 20 / 15 & 20 / 9 & 20 / 11 & 101  &  180  &  87 K  \\
\bottomrule
\end{tabular}}
\vspace{-4mm}
\end{center}

\end{table*}

\section{Proposed Benchmark}
\label{sec:benchmark}

The proposed benchmark attempts to provide the community an evaluation platform that assesses trackers under various scenarios closer to real applications efficiently and effectively.
To this end, we follow the principles of efficiency, diversity, balance, and inheritance to construct the benchmark.
For efficiency, we collect sequences with a high density of challenging variations and avoid long-range frames without changes.
Taking both appearance and variation patterns into consideration, we summarize 9 different types of scenarios and multiple diverse sub-scenarios to ensure the diversity principle.
For each scenario, we collect 20 sequences with a similar average frame length to strike a balance among different scenarios.
As existing benchmarks have their own characteristics and challenging factors despite the saturated performance, we select the most challenging and effective sequences from them based on a proposed quality metric.
In the following, we first describe the evaluated methods and then present the collected data.

\subsection{Evaluated Methods}
Considering most deep trackers are built on one of the Siamese, classifier+IoUNet, and transformer tracking frameworks, we select representative trackers for each framework and three other representative methods with different characteristics.
Table~\ref{Tab:methods} shows the detailed features of the evaluated tracking algorithms.
Although the source codes of these methods are all publicly available, their training settings, especially for the training and validation data are not consistent, which may influence the performance significantly.
For fair comparisons, we re-train all these methods using the same training settings as much as possible.
Specifically, we perform offline training on the widely used training datasets of LaSOT~\cite{LaSOT}, TNet~\cite{TrackingNet}, and Got-10K~\cite{GOT10k}, which are specifically designed for tracking and cover various categories and scenarios.
For the hyper-parameters related to online tracking, we pick them based on the performance over the validation set of Got-10K.

\subsection{Data Collection}
\label{Sec:data-collection}
The proposed ITB is supposed to be efficient and effective with a reduced total frame number but covers diverse application-oriented scenarios compared to existing datasets.
To this end, we propose a quality metric to provide an objective quantitative description for selecting effective and efficient sequences and define 9 tracking scenarios based on both appearance and variation patterns to ensure diversity.

\vspace{1mm}
\noindent\textbf{Metric of sequence quality.}
The quality metric should be able to describe how effective a sequence is in evaluating tracking methods.
The effectiveness of a test sequence lies in how challenging it is and how well it can distinguish the performance of different trackers.
In addition, an effective test sequence should have a high density of challenging variations and avoid long duration of easy frames to ensure its efficiency for evaluation.
Therefore, the quality score of a test sequence is related to challenge degree, discriminative ability, and variation density.

In light of that the tracking performance of multiple methods over a sequence also reflects the performance of this sequence for providing effective evaluation, we define the quality metric based on the quantitative results of a variety of state-of-the-art trackers.
\begin{figure}
    \centering
    \includegraphics[width=0.99\linewidth]{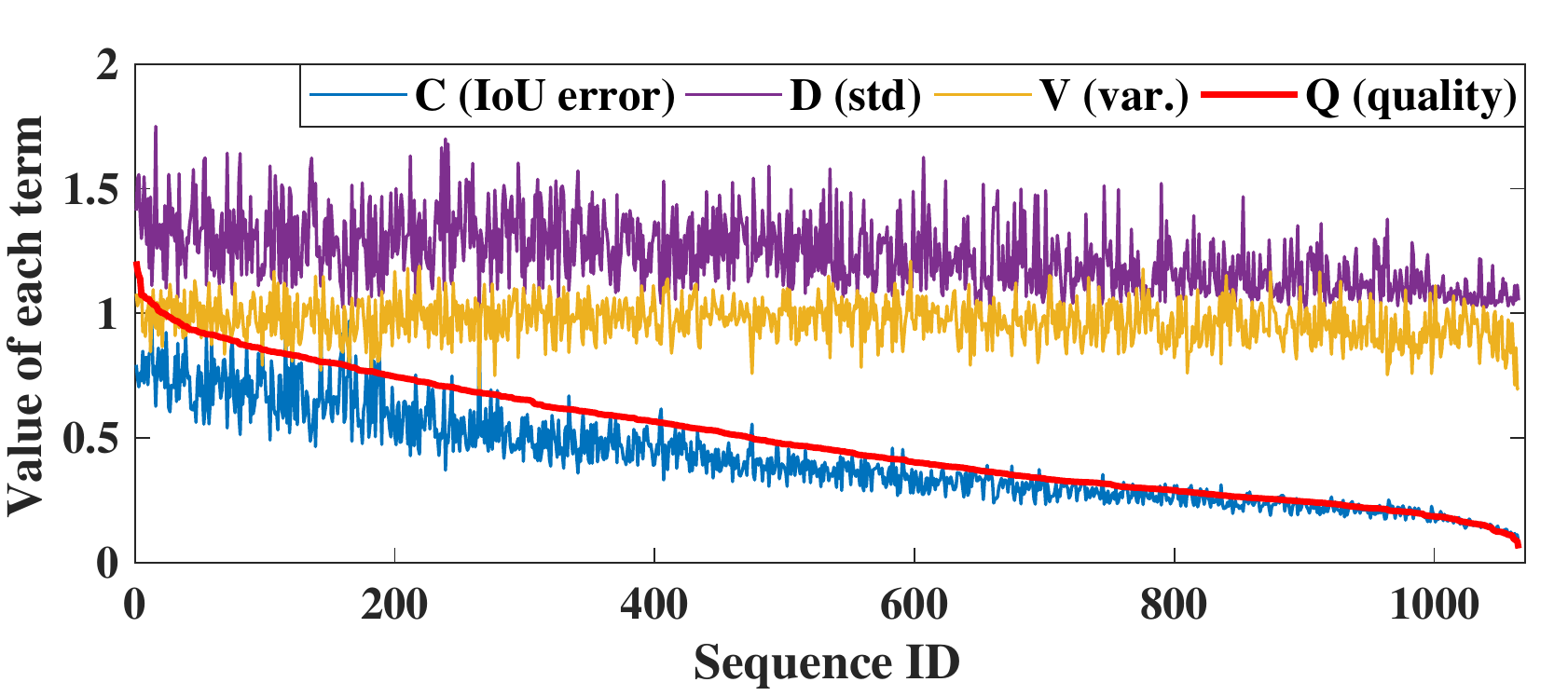}
    \caption{\textbf{Distributions of each term on the 1065 sequences from existing datasets.} C, D, V, Q denote challenge degree (based on IoU error), discriminative ability (based on std of IoU score), variation density, and quality score of a sequence. It shows that the sequence with a higher challenge degree, more distinguishable performance of different methods, and a higher variation density obtains a higher quality score, which is more effective in evaluating tracking methods.}
    \label{fig:quality-score}
    \vspace{-4mm}
\end{figure}

\begin{figure*}[!htb]  
\renewcommand\arraystretch{1.0}
\setlength\tabcolsep{1pt}
\resizebox{1.0\linewidth}{!}{
\begin{tabular}{ccccccccc}
    \textbf{Human body}  & \textbf{Human part}    & \textbf{Animal}   & \textbf{Sign and logo}   & \textbf{Vehicle }   & \textbf{3D-object }     & \textbf{Sport ball}    & \textbf{UAV}   & \textbf{Cartoon} \\ 
            \includegraphics[width=0.11\linewidth]{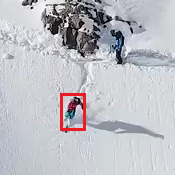} & 
            \includegraphics[width=0.11\linewidth]{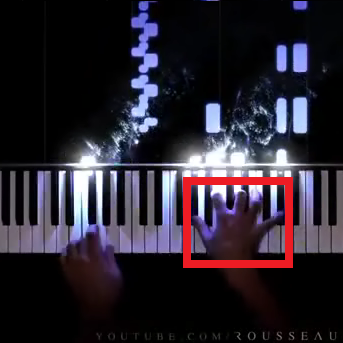} &
            \includegraphics[width=0.11\linewidth]{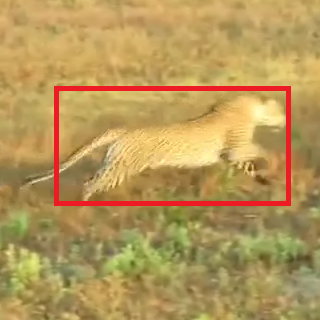} &
            \includegraphics[width=0.11\linewidth]{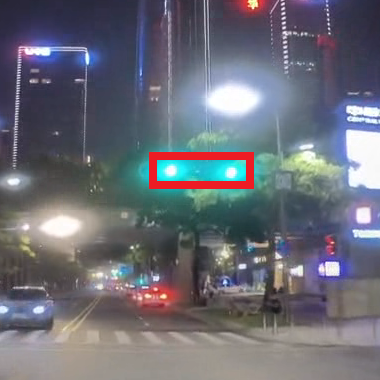} &
            \includegraphics[width=0.11\linewidth]{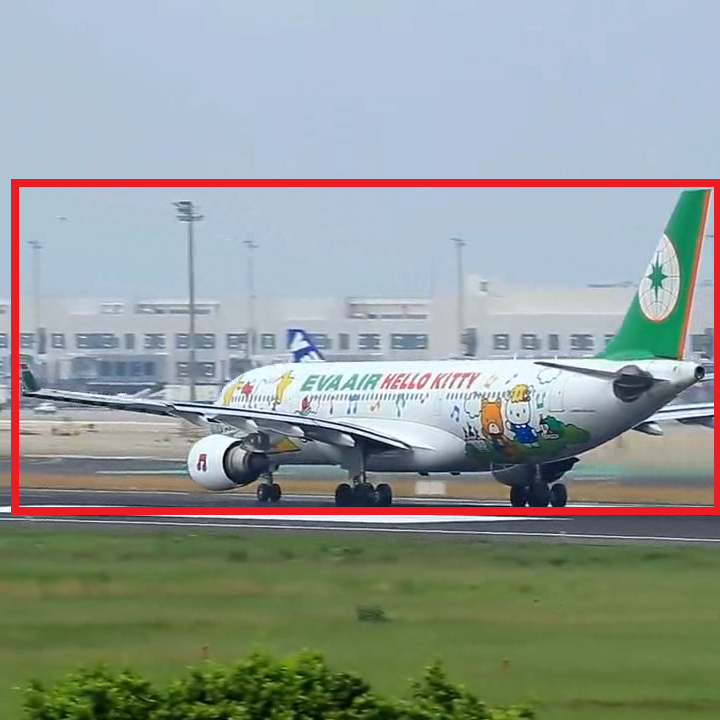} &
            \includegraphics[width=0.11\linewidth]{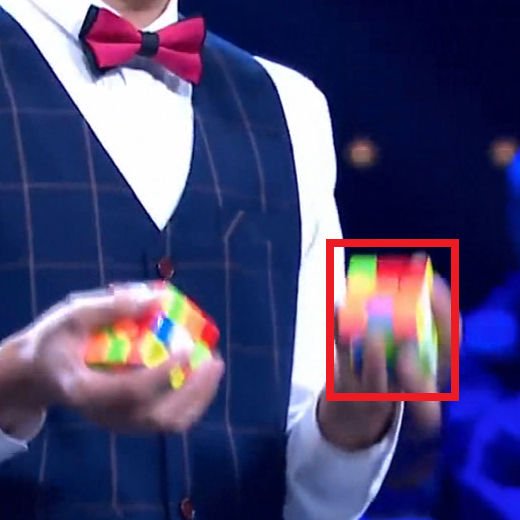} &
            \includegraphics[width=0.11\linewidth]{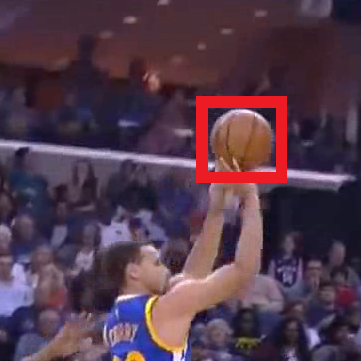} &
            \includegraphics[width=0.11\linewidth]{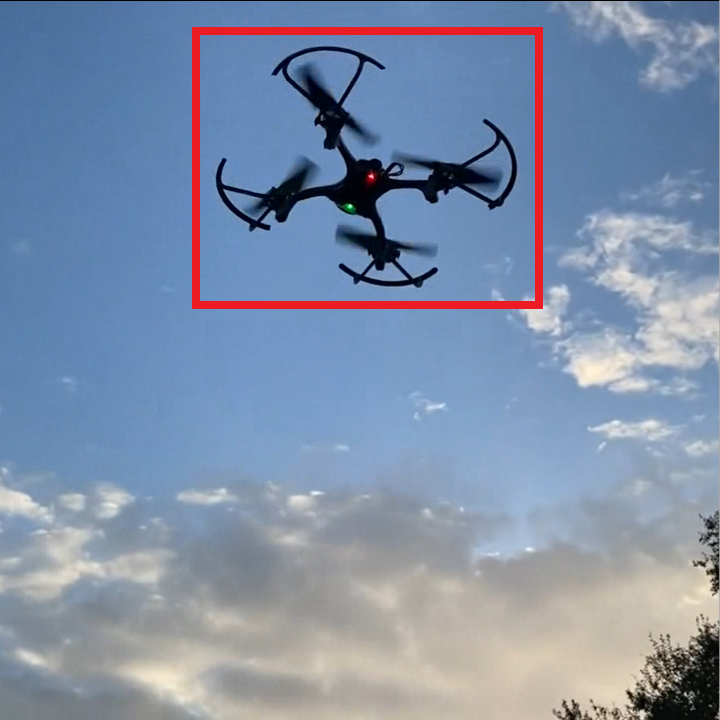} &
            \includegraphics[width=0.11\linewidth]{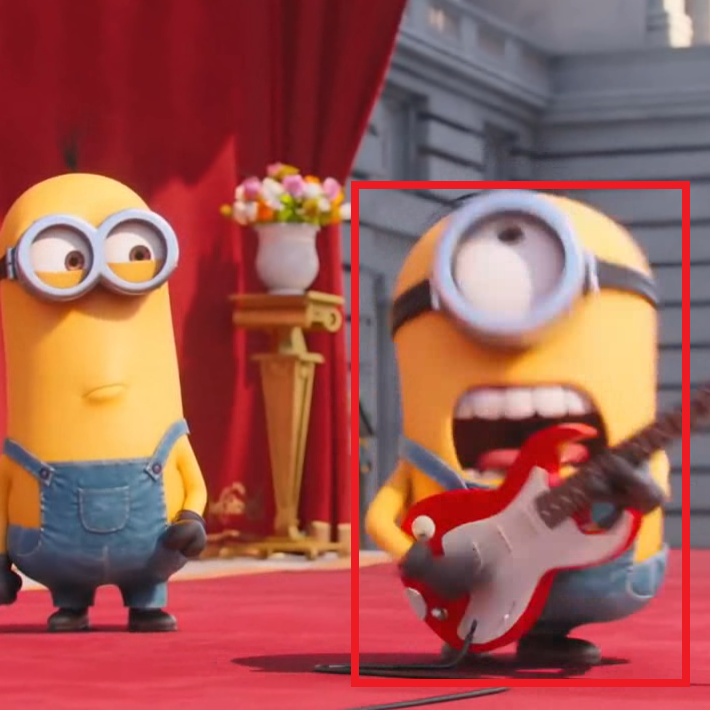} \\ 
            \includegraphics[width=0.11\linewidth]{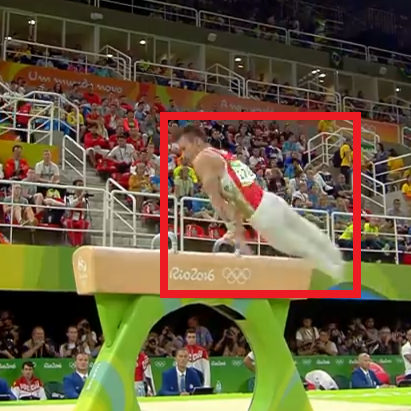} & 
            \includegraphics[width=0.11\linewidth]{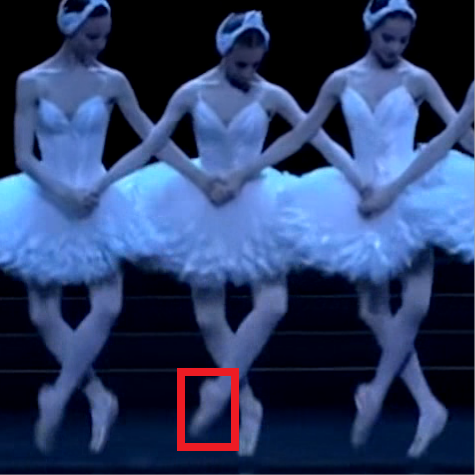} &
            \includegraphics[width=0.11\linewidth]{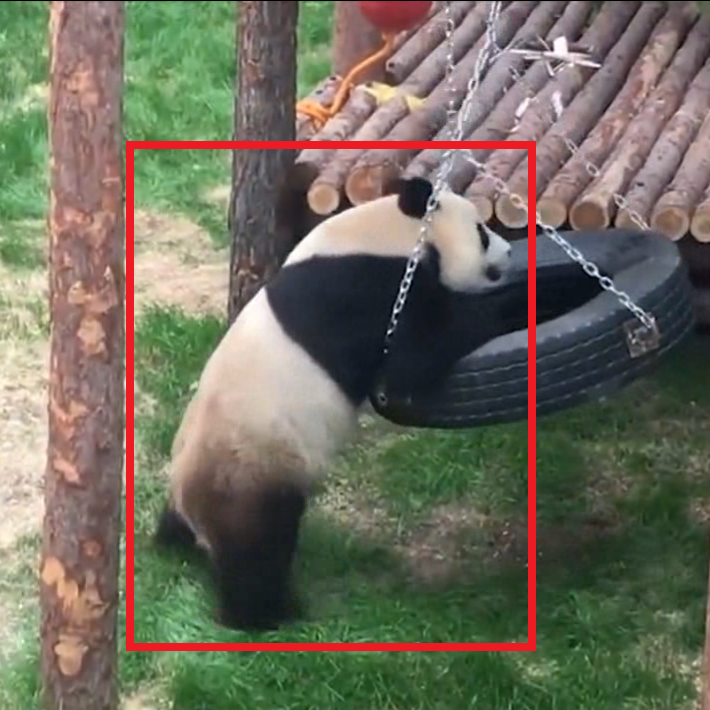} &
            \includegraphics[width=0.11\linewidth]{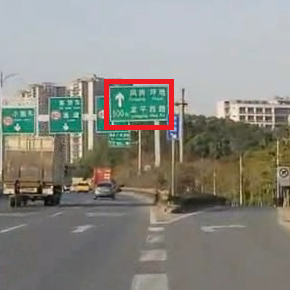} &
            \includegraphics[width=0.11\linewidth]{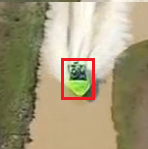} &
            \includegraphics[width=0.11\linewidth]{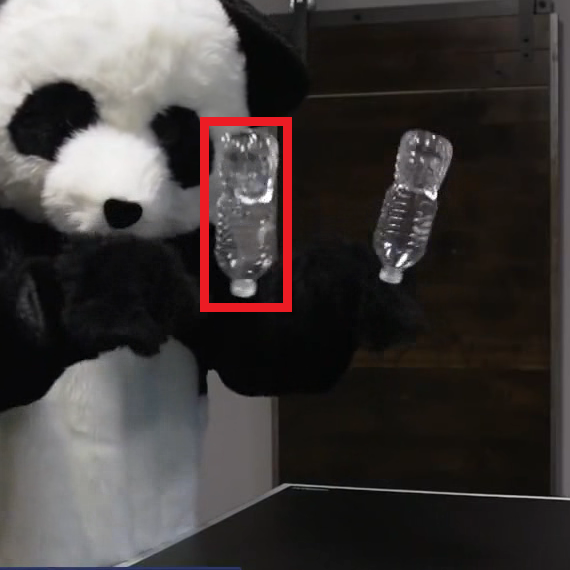} &
            \includegraphics[width=0.11\linewidth]{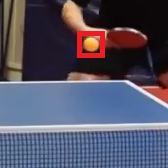} &
            \includegraphics[width=0.11\linewidth]{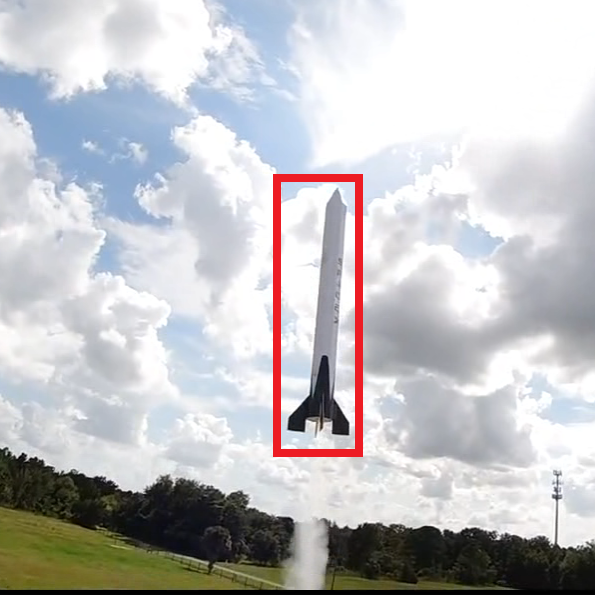} &
            \includegraphics[width=0.11\linewidth]{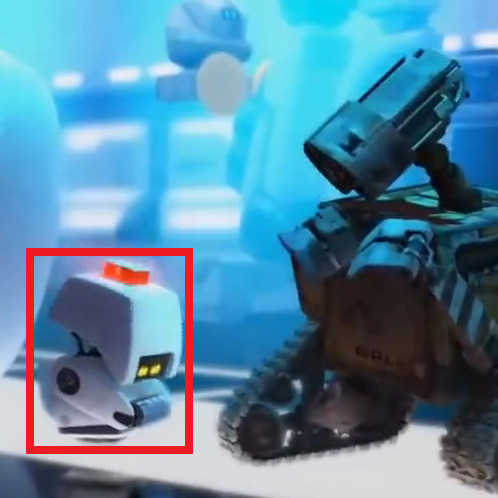} \\ 
            \includegraphics[width=0.11\linewidth]{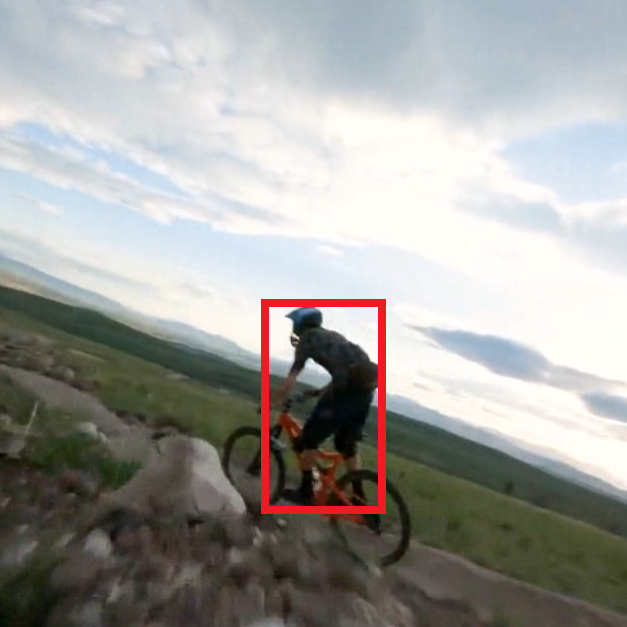} & 
            \includegraphics[width=0.11\linewidth]{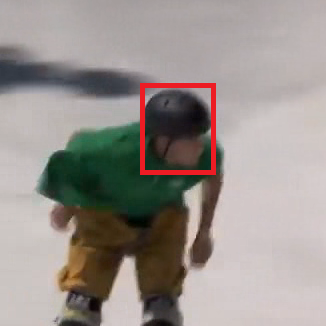} &
            \includegraphics[width=0.11\linewidth]{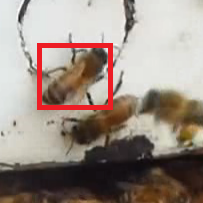} &
            \includegraphics[width=0.11\linewidth]{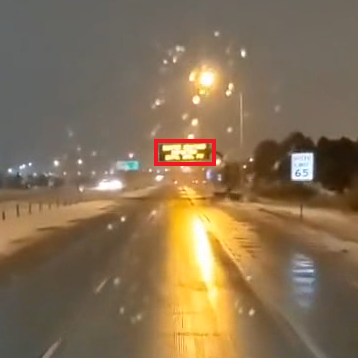} &
            \includegraphics[width=0.11\linewidth]{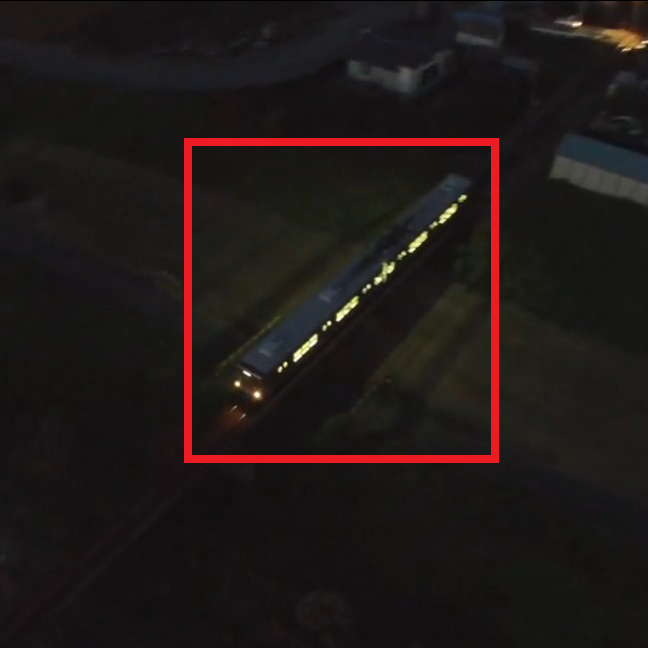} &
            \includegraphics[width=0.11\linewidth]{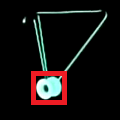} &
            \includegraphics[width=0.11\linewidth]{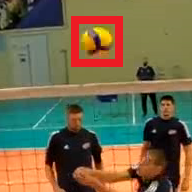} &
            \includegraphics[width=0.11\linewidth]{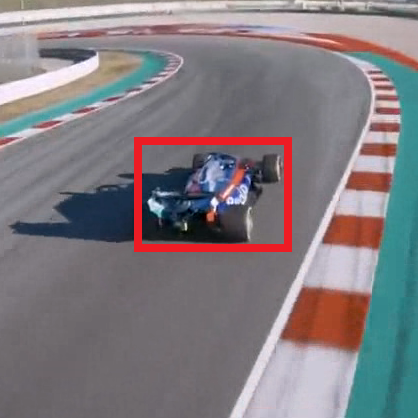} &
            \includegraphics[width=0.11\linewidth]{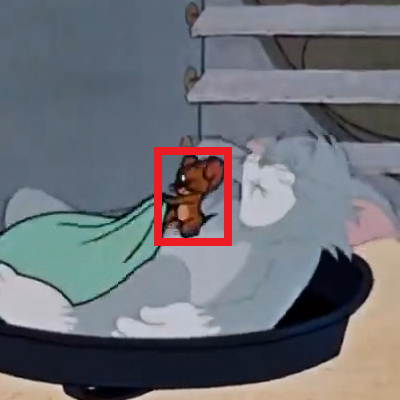} \\ 
            \includegraphics[width=0.11\linewidth]{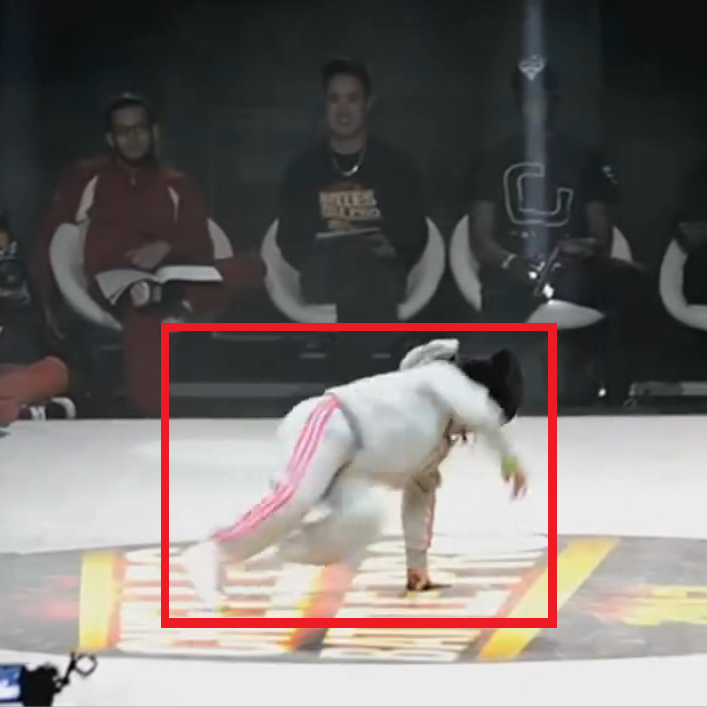} & 
            \includegraphics[width=0.11\linewidth]{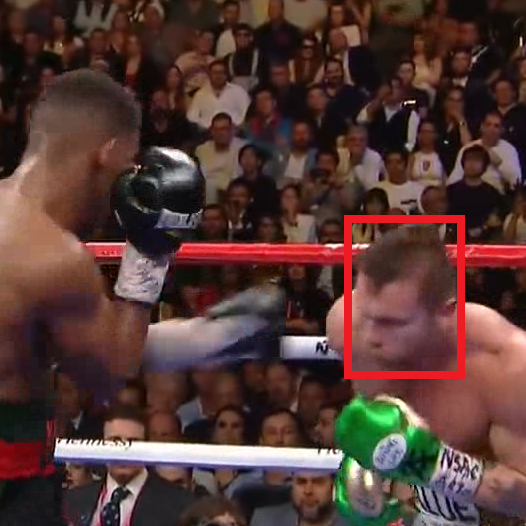} &
            \includegraphics[width=0.11\linewidth]{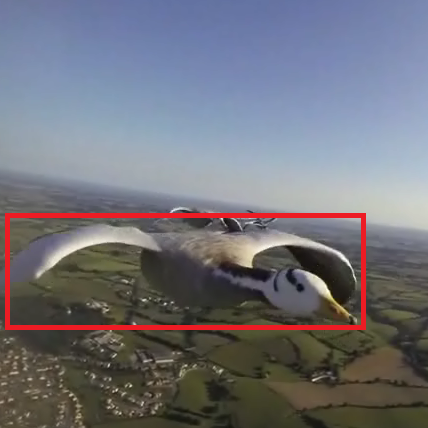} &
            \includegraphics[width=0.11\linewidth]{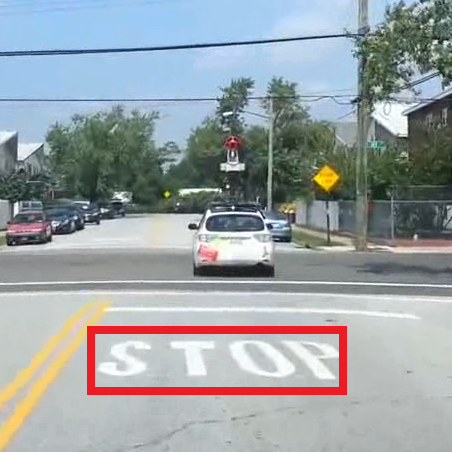} &
            \includegraphics[width=0.11\linewidth]{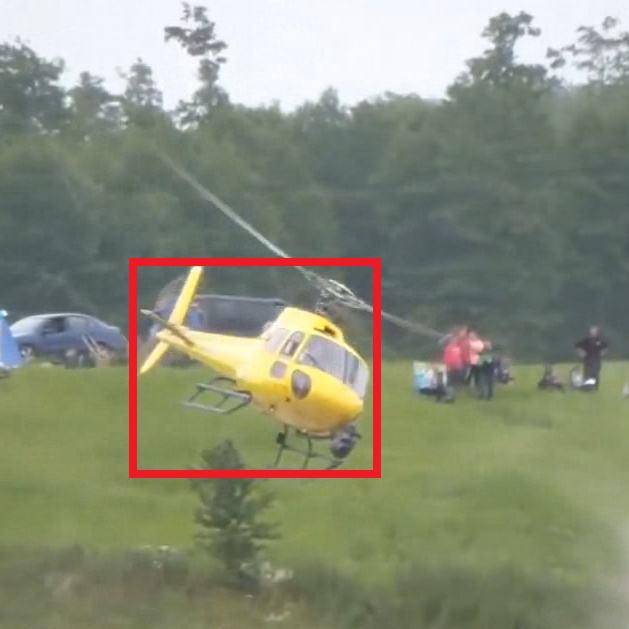} &
            \includegraphics[width=0.11\linewidth]{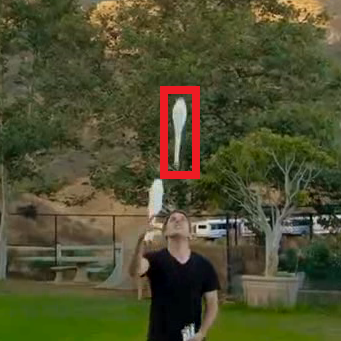} &
            \includegraphics[width=0.11\linewidth]{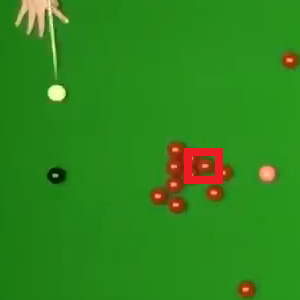} &
            \includegraphics[width=0.11\linewidth]{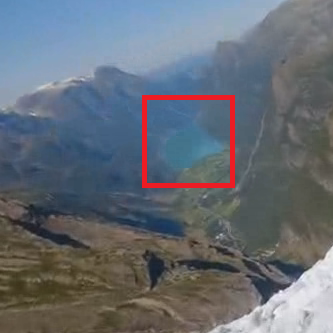} &
            \includegraphics[width=0.11\linewidth]{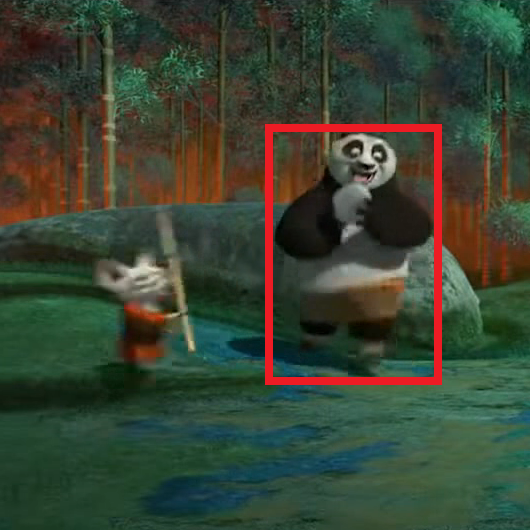} \\ 
\end{tabular} }
\vspace{-2mm}
	\caption {\textbf{ Screenshots of some sequences from the 9 scenarios.} The sequences in every column belong to the same category whose name is given in the top. The target object in each sequence is annotated with a red bounding box. It shows that the target objects are with a wide range of appearance and variation patterns which differ significantly.}
	\label{fig:screenshot-seqs}
	\vspace{-2mm}
\end{figure*}


    		
    



%
The Intersection over Union (IoU) score~\cite{OTB2013} well demonstrates the tracking performance, meanwhile, it also shows the difficulty of the test sequence.
Since the lower IoU score, the bigger the challenge, we formulate the challenge degree of a sequence as
\begin{equation}
    C=1-\mu_{i,j}(S_{i,j}),
\end{equation}
where $S_{i,j},(i\in [1,N], j\in [1,M])$ is a two-dimensional matrix denoting the IoU scores of $N$ methods on a sequence with $M$ frames, $\mu_{ij}(\cdot)$ computes the mean value of $M_{i,j}$.
%
As the standard deviation (std) of IoU scores shows how diverse the performance between different methods are, we use it to measure the discriminative ability of a sequence, which is defined as
\begin{equation}
    D=\exp(\eta \sigma_i( \overline{S}_{i})),
\end{equation}
where $\overline{S}_{i}$ is the mean IoU score of the $i$-th method over all the frames, $\sigma_i$ computes the IoU std of all the methods, $\eta$ is a scale factor for scaling the discriminative ability score to a proper range, and $\exp(\cdot)$ denotes the exponential function taking the base of natural logarithms as the base value.
%
We use the $\exp(\cdot)$ function to map the std values in [0,1] to values larger than 1.
Note that challenging variations usually cause changes of tracking performance between adjacent frames.
%
Based on this observation, we compute the variation density score based on the changes of IoU scores in adjacent frames and frame length, which is formulated as
\begin{equation}
    V_{ar}= \mathcal{N}_{a,b}(\log( \sum_{i,j}{S_{i,j+1}-S_{i,j}} )) / \mathcal{N}_{a,b}(\log(M)),
\end{equation}
where $M$ is the frame number of the sequence, and $\mathcal{N}_{a,b}$ is a normalization function that normalizes the input values to the range of [a,b] in a min-max manner.
Using these terms, the quality score is defined as
\begin{equation}
        Q(S)=C(S)*D(S)*V_{ar}(S).
    \label{eq:quality-score}
\end{equation}
Fig.~\ref{fig:quality-score} shows the values of each term corresponding to all the 1065 sequences from the existing datasets of OTB100, NFS, UAV123, LaSOT, VisDrone (including the train and validation sets), and NUS-PRO.
Instead of picking test sequences manually in a subjective way, Eq.~\ref{eq:quality-score} provides a quantitative description of sequence quality for collecting data more effectively and objectively.


\vspace{1mm}
\noindent\textbf{Diversity of scenarios.}
In the context of visual tracking, the target object can be any category and its variations can be diverse and complicated.
Therefore, to provide a comprehensive evaluation, it is crucial for the test dataset to be with high diversity in both terms of target object category and variation patterns.
In addition, it is more meaningful to see how a tracker performs in real applications, such as tracking the human body or animal for trajectory analysis, tracking part of the human body for action and posture recognition, and tracking traffic signs or vehicles for an automatic drive.

Taking all factors into consideration, we define 9 typical scenario categories which are \textit{human body}, \textit{human part}, \textit{animal}, \textit{vehicle}, \textit{signs and logos}, \textit{sport balls}, \textit{small rigid objects} (such as dolls and boxes), \textit{UAV} (including the videos with a UAV target or captured in the UAV view), and \textit{Cartoon}.
To ensure the diversity and comprehensiveness for each category, we collect typical subcategories with different appearance or motion patterns as diverse as possible.
The appearance difference between subcategories includes different finer classes (such as hand, head, and foot in the \textit{human part} category) and the variation difference usually lies in different applications or activities (such as a hand playing piano, and a hand showing sign language).
We select sequences with representative appearance and variation patterns as best as we can and ensure that every category contains at least 10 different appearance and variation patterns.
For instance, the \textit{human body} scenario includes the subcategories of the pedestrian with slight variations, runner with deformation and distractors, dancer and gymnast with more complicated variation patterns involving out-of-plant and in-plant rotations, deformation, occlusion \etc.

\begin{figure*}
  \centering
  \begin{subfigure}{0.476\linewidth}
    \includegraphics[width=1\linewidth]{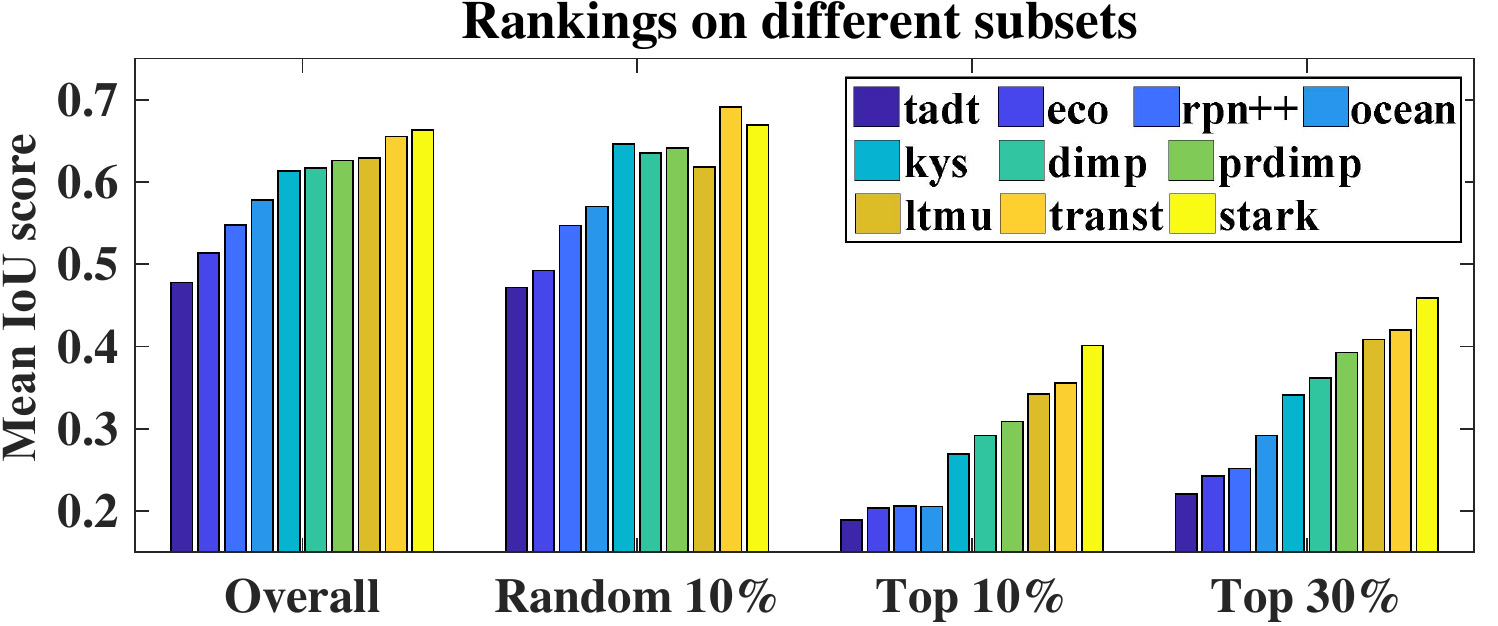}
    \caption{Performance ranking of the 10 base trackers.}
    \label{fig:ranking-method10}
  \end{subfigure}
  \hfill
  \begin{subfigure}{0.47
  \linewidth}
    \includegraphics[width=1\linewidth]{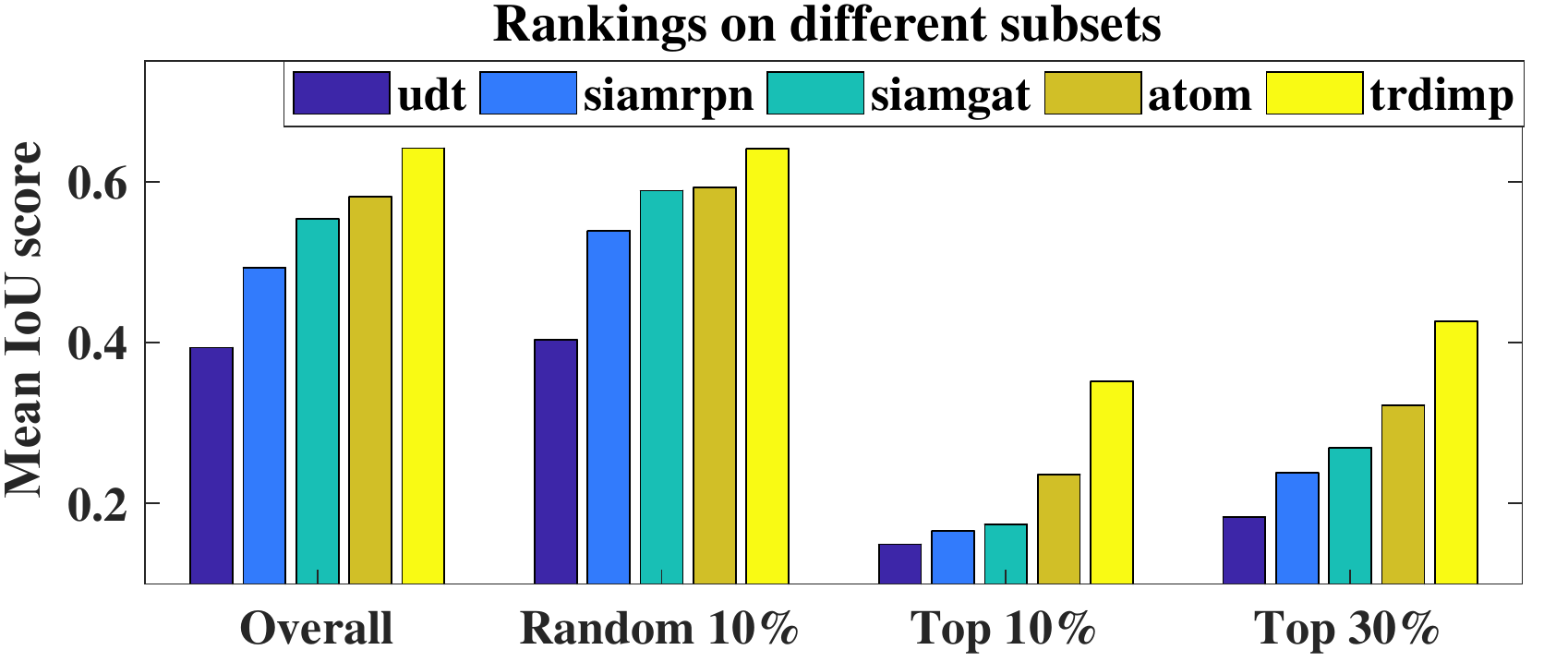}
    \caption{Performance ranking of the 5 validation trackers.}
    \label{fig:ranking-method5}
  \end{subfigure}
  \vspace{-1mm}
  \caption{\textbf{ Performance comparison on different subsets.} The figure shows the performance on the subsets of all the sequences (Overall), randomly selected 10\% sequences (Random 10\%), top 10\% sequences (Top 10\%) and top 30\% sequences (Top 30\%) with the highest quality score. (a) and (b) present the performance of the 10 base trackers whose performance is used for sequence quality computation, and other 5 methods, respectively. It demonstrates that the sequences with a higher quality score are more effective for evaluation in terms of challenge degree, discriminative ability, and efficiency.}
  \label{fig:ranking-comparison}
  \vspace{-2mm}
\end{figure*}

Table~\ref{Tab:datasets} shows the sequence and subcategory numbers of each category, and compares the proposed benchmark against existing datasets.
It shows that the defined 9 scenarios can cover almost all the sequences in existing datasets and the proposed dataset contains the most subcategories with a total number of 101.
Fig.~\ref{fig:screenshot-seqs} presents the screenshots of some sequence from every scenario.
The target objects in these sequences differ significantly in terms of appearance and variation patterns.

\vspace{1mm}
\noindent\textbf{Data source and annotations.}
Since every existing tracking dataset has its own characteristic and contains some sequences that are still challenging to state-of-the-art methods, we select the top 10 percent of sequences with the highest quality score (defined in Eq.~\ref{eq:quality-score}) from the datasets of OTB100~\cite{OTB2015}, NFS~\cite{NFS}, UAV123~\cite{UAV123}, LaSOT~\cite{LaSOT}, NUS-PRO~\cite{NUS-PRO}, and VisDrone~\cite{VisDrone}.
After that, we remove redundant sequences belonging to the same subcategory with similar appearance and variation patterns to ensure each subcategory only includes one sequence.
In addition, we collect sequences from the youtube website by downloading public videos and carefully selecting clips from them.
We first generate a large number of video clips for each scenario and then add the best ones of them based on the quality score to the dataset and make sure each scenario includes 20 sequences.
We analyze the videos carefully and avoid selecting videos that contain personal information.

\subsection{Evaluation Protocol}
\label{sec:protocol}
The inconsistent training data used by different methods and the finetuning step directly on the testing datasets undermine the fairness of comparison.
The training and validation sets are not provided in earlier datasets but several recent datasets~\cite{TrackingNet, LaSOT, GOT10k} provide training sets.
However, existing methods still use multiple training sets, which may be because more data from diverse datasets will contribute to better performance.

To ensure fair comparisons and be compatible with the settings of previous methods, we take the integration of the training sets from the TNet, LaSOT, and Got10-K datasets as the training data.
In addition, we use the validation set of Got10-K to select trained models in each epoch and finetune hyper-parameters.
For testing, we use the 3-pass evaluation (running a tracker on all the sequences 3 times and using the mean score as the results), considering the performance of some methods are not stable.
The mean IoU score (mIoU) is used to rank methods.

\section{Experiments}
\label{sec:formatting}

In this section, we first demonstrate the effectiveness of the proposed sequence quality metric.
We then present the overall performance of the 15 state-of-the-art methods on the proposed benchmark and compare it with that on other datasets to show its effectiveness in providing comprehensive and efficient evaluation.
Furthermore, we show the performance over every scenario to identify the state-of-the-art methods under different scenarios.
Finally, we discuss the new challenges in real applications that badly influence tracking performance but have not been well studied.

In all evaluations, we use the source codes and the default training and testing settings except for the training data and the hyper-parameters used in the test stage.
We use the same training data (training sets of TNet, LaSOT, and Got10-K) for the methods that require offline training, except for the UDT and LTMU methods which use the default training data since they need to be trained on samples from successive frames and long-term sequences.
We use the validation set of Got-10K for checkpoint selection and hyper-parameters finetuning, and use the same checkpoint and hyper-parameters for all evaluations.
The speeds shown in Table~\ref{Tab:methods} are tested on a PC with a Xeon CPU (2.30GHz) and a Tesla-V100 GPU card.
More results and dataset information are available in the supplementary material.


\begin{table*}[t]
\begin{center}
\caption{\textbf{ Performance of 10 state-of-the-art methods on ITB and all benchmarks, and the comparisons between them in terms of challenge degree, discriminant ability, and evaluation efficiency.} 
The left part shows the mIoU scores of each method on each benchmark and the right part compares the stats of mean mIoU, normalized std (NStd, computed by dividing std by mean mIoU), and test time (minute) for each dataset.
}
\label{Tab:overall_results}
\vspace{-2mm}
\scriptsize
\renewcommand\arraystretch{1.1}
\setlength\tabcolsep{4pt}
 \resizebox{0.98\linewidth}{!}{
\begin{tabular}{l| cccccccccc|ccr}
\toprule      
          \multicolumn{1}{r|}{Trackers }&\multicolumn{3}{c|}{Siamese} &\multicolumn{4}{c|}{Classifier+IoUNet} &\multicolumn{3}{c|}{Transformer} &\multirow{2}{*}{\tabincell{c}{mean\\mIoU }$\downarrow$} &\multirow{2}{*}{\tabincell{c}{NStd\\mIoU }$\uparrow$}&\multirow{2}{*}{\tabincell{c}{Test\\time}$\downarrow$}\\
Datasets & RPN++ & Ocean & \multicolumn{1}{c|}{GAT} & ATOM  & DiMP  & PrDiMP  & \multicolumn{1}{c|}{KYS}   & TrDiMP &TransT &Stark & & &\\
\midrule
OTB100~\cite{OTB2015}  & 66.6 & 65.6 & 65.3 & 65.6 & 67.4 & 68.7 & 69.4 & 69.7 & 67.0 & 69.0 & 67.4  & 2.40 & 41 m \\
NUS-PRO~\cite{NUS-PRO} & 53.1 & 56.7 & 52.7 & 56.6 & 59.1 & 59.2 & 60.5 & 60.2 & 62.8 & 62.4 & 58.3  & 5.59& 94 m \\
UAV123~\cite{UAV123}  & 57.8 & 58.3 & 59.1 & 63.2 & 64.8 & 66.7 & 63.7 & 65.4 & 65.5 & 68.6 & 63.3  & 5.42& 78 m \\
VisDrone~\cite{VisDrone}& 57.2 & 63.1 & 54.5 & 57.4 & 61.1 & 54.8 & 61.6 & 62.5 & 60.7 & 60.2 & 59.3  & 4.94& 53 m \\
NFS(30)~\cite{NFS} & 49.7 & 54.3 & 52.6 & 58.8 & 62.4 & 62.9 & 62.9 & 64.1 & 63.6 & 65.6 & 59.7  & 8.46& 33 m\\
TNet~\cite{TrackingNet}    & 69.0 & 69.2 & 71.7 & 72.5 & 74.9 & 75.8 & 74.1 & 78.3 & 80.7 & 81.1 & 74.7  & 5.33& 156 m\\
Got10K~\cite{GOT10k}  & 48.8 & 59.9 & 53.8 & 56.1 & 61.4 & 64.6 & 62.1 & 68.2 & 70.2 & 68.8 & 61.2  & 9.91& 16 m\\
LaSOT~\cite{LaSOT}   & 49.2 & 55.2 & 51.0 & 51.0 & 57.4 & 60.8 & 53.4 & 62.4 & 65.5 & 66.3 & 57.2  & 9.42& 476 m\\ \hline
\textbf{ITB (ours)}& 44.1 & 47.7 & 44.9 & 47.2 & 53.7 & 54.4 & 52.0 & 56.1 & 54.7 & 57.6 & 51.2 & 8.44& 60 m \\
\bottomrule
\end{tabular}}
\end{center}
\vspace{-4mm}
\end{table*}

\subsection{Effectiveness of the Sequence Quality Metric}
To demonstrate the correctness of the proposed metric in measuring the challenge degree and discriminative ability of a sequence, we compare the performance rankings of 15 trackers on the sets of the top 10\% and top 30\% sequences with the highest quality score with those on all the sequences and the set of 10\% randomly selected sequences from the test sequences of the OTB100, NFS, UAV123, LaSOT, VisDrone, and NUS-PRO datasets (1065 sequences in total).
%
We select the 15 representative and state-of-the-art tracking methods with various tracking frameworks (given in Table~\ref{Tab:methods}).
For fair and objective evaluations, we split the 15 trackers into a base part with 10 trackers whose results are used to compute the quality score of each sequence, and a test part with the remaining 5 trackers, and ensure that each part covers all the tracking frameworks in Table~\ref{Tab:methods}.

Fig.~\ref{fig:ranking-comparison} shows the AUC scores of all these trackers on different sets of sequences.
We draw the following observations from these results.
First, the AUC scores on the top 10\% and top 30\% sequences with the highest quality score are much lower than those on the overall sequences and on the randomly selected 10\% sequences.
In addition, the AUC scores on the top 10\% sequences are also lower than those of the top 30\% sequences.
This shows that the sequences with high-quality scores are more challenging and the top 10\% sequences are more difficult than those in the top 10\% to top 30\%, which demonstrates that the quality score metric well measures the challenge degree of a sequence.
Second, the top 10\% to top 30\% sequences provide more distinguishable performance compared with that on the overall sequences.
This indicates that the quality score metric can identify the sequences that are more effective in distinguishing the performance of different methods.
Third, the rankings on the selected subsets remain the same as those on the overall sequences, which indicates that the selected sequences are the key ones for ranking the performance, while others may be redundant and less effective.
%
Fourth, we have the same observations from the results on both the base trackers and the 5 test trackers. which shows that the quality score computed based on the representative trackers generalizes well to other tracking methods.

\begin{figure}
    \centering
    \includegraphics[width=0.92\linewidth]{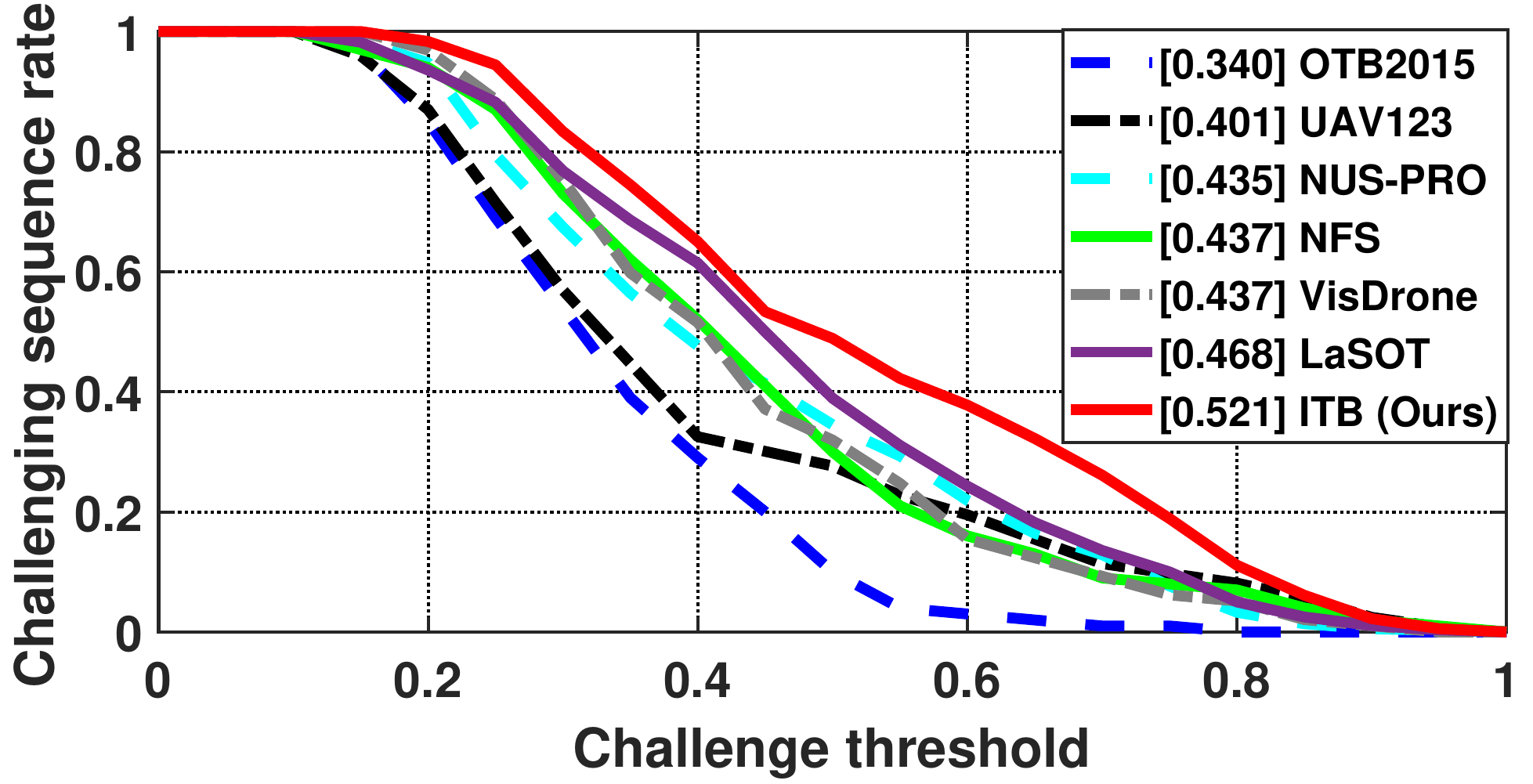}
    \vspace{-1mm}
    \caption{\textbf{Challenge plots of the sequences from the 7 benchmarks.} The X and Y coordinates correspond to the mIoU error (1-mIoU) threshold for judging whether a sequence is challenging or not, and the ratio of challenge sequences to all sequences in a benchmark. The number in the legend is the area under the curve which indicates the challenge level in an overall view. }
    \label{fig:challenge-distribution}
    \vspace{-4mm}
\end{figure}

\begin{figure*}
    \centering
    \includegraphics[width=0.99\linewidth]{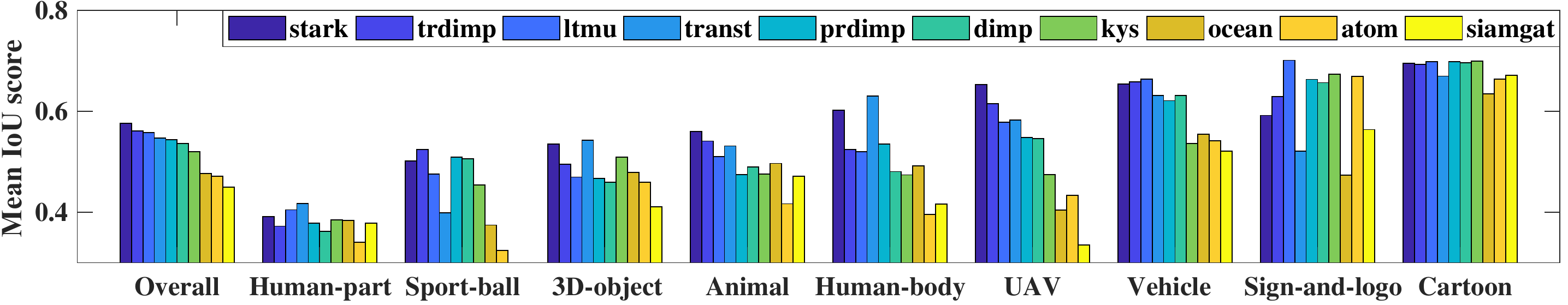}
    \vspace{-1mm}
    \caption{\textbf{Performance of the top 10 ranking methods on the subsets of each scenario.} It shows that the performance of the methods varies greatly over different scenarios, providing a comprehensive and in-depth evaluation.}
    \label{fig:scenario-performance}
    \vspace{-4mm}
\end{figure*}

\subsection{Overall Performance}
\label{sec:overall_perf}
To analyze the overall performance of state-of-the-art methods on the ITB benchmark and the effectiveness of the proposed benchmark, we conduct experimental evaluations of 10 representative methods given in Table~\ref{Tab:methods} using the protocol in Section~\ref{sec:protocol} on 8 existing benchmarks and ITB.
Table~\ref{Tab:overall_results} shows the mIoU score (\%) of the 10 methods on all these benchmarks.
On ITB, the transformer-based methods achieve the best performance attributed to the advanced representation capacity and the Classifier+IoUNet trackers rank second benefited from the online learning scheme compared with the Siamese-based approaches.
The performance ranking of each kind of trackers remains the same on all the 8 existing benchmarks.
Among the transformer-based trackers, Stark achieves the best performance and ranks first on ITB and other 4 out of 8 benchmarks.
The performance ranking of other methods on ITB is similar to the overall ranking on the 8 existing benchmarks, which shows that ITB provides comprehensive evaluation.
The right part of Table~\ref{Tab:overall_results} shows the statistics of the results on each dataset including the mean, std, and normalized std (NStd) of the mIoU scores over all the methods, and test time.
The ITB obtains the lowest mean mIoU score, a high NStd score, and a short test time, which demonstrates that ITB leaves more space for trackers to improve, provides discriminative performance of different methods, and enables efficient evaluations.

Fig.~\ref{fig:challenge-distribution} shows the distribution of challenging sequences from evaluated benchmarks.
We recognize a sequence as a challenging one when the mean mIoU error (1-mIoU) of the 15 methods on it, is larger than a given threshold.
The plots in Fig.~\ref{fig:challenge-distribution} show that our ITB is constructed with more balanced sequences with different challenge degrees and the highest overall challenge level, which contributes to effective evaluation shown in Table~\ref{Tab:overall_results}.

\subsection{Scenario-Based Analysis and New Challenges}
\label{sec:scenario-analysis}
With the experimental results on the constructed subsets of the 9 scenarios, we analyze the performance of the top 10 ranking trackers to study their effectiveness in tracking objects with a specific category and typical appearance variations.
Fig.~\ref{fig:scenario-performance} shows the mIoU scores of all the trackers on each set.
Overall, both the performance over the different scenarios and the performance of different methods vary greatly.
Among these scenarios, \textit{Human-part} is the most challenging one with the lowest average mIoU score less than 0.4.
The challenging factors in the \textit{Human-part} scenario include uncertain appearance patterns (\eg hand with different gestures), inconspicuous and incomplete boundaries, and similar distractors (\eg tracking one foot is easily influenced by the other foot).
Although the transformer models with advanced representations obtain relatively better performance, there is much room to improve.
\textit{Sport-ball} is the second most challenging scenario but with a more diverse performance of different trackers.
The challenging factors include tiny object with less appearance information, fast motion with irregular and sudden speed changes, motion blur, and occlusion (\eg the sequences with the target of soccer, golf, or pingpang.)
The tracking methods (\eg Stark and DiMP) with powerful representations and online learning schemes achieve better performance, while the Siamese-based approaches using a target motion prior with a Gaussian-like weight map do not perform well.
The overall performance on the scenarios of \textit{Vehicle}, \textit{Sign-and-logo}, and \textit{Cartoon} are relative better.
This is because the target objects in the subsets of \textit{Vehicle} and \textit{Sign-and-logo} are rigid and have a stable appearance model, while the subset of \textit{Cartoon} enables rich color features.
Due to space limitations, we analyze the performance on other scenarios in the supplementary material.

Based on the scenario-based performance of 10 state-of-the-art trackers, we find the following challenges that widely exist in tracking scenarios and have not been well studied.
 \textbf{1) Abrupt appearance change}. Targets may change rapidly to be a totally different appearance, which is usually happened in the open-world under the cases where the target object with an irregular 3D structure goes through view change or out-of-plane rotation.
 \textbf{2) Noisy initialization}. The bounding box annotation form usually makes the initial target patch contain background context, which badly degrades tracking performance, especially when the background context fills a large part of the target patch.
 \textbf{3) Totally similar-looking distractor}. The distractors with an extremely similar appearance appear commonly in numerous scenarios, \eg hand/foot tracking, human tracking with the same uniform.

\subsection{Discussion and Limitations}
Although the proposed ITB includes 9 typical scenarios and 101 diverse sub-scenarios, there are still other representative and challenging scenarios to be involved.
As the quality metric of sequences is defined based on the performance of the 10 state-of-the-art methods, the qualities of sequences need to be updated when more and more new state-of-the-art methods are proposed.
%

\section{Concluding Remarks}
In this work, we introduce an informative tracking benchmark covering all typical challenging scenarios with dense appearance variations for enabling comprehensive and efficient performance evaluation.
We first develop a quality assessment mechanism for sequence selection based on the performance of 10 state-of-the-art trackers taking challenging level, discriminative strength, and density of appearance into consideration.
With the mechanism, we select the most high-quality sequences from existing benchmarks and collect informative sequences to ensure diversity and balance between different scenarios.
The proposed ITB allows comprehensive performance evaluation using 93\% less time while remaining similar performance ranking and leaving much space for tracking methods to improve.
The experimental results on ITB demonstrate that performance evaluation on sequences from diverse scenarios benefits a better understanding of the state-of-the-art methods and helps reveal new tracking challenges.

{\small
\bibliographystyle{ieee_fullname}
\bibliography{tracking}
}

\end{document}